\documentclass{elsarticle}
\usepackage{amssymb}

\journal{arXiv}

\usepackage{subfiles}
\usepackage{blindtext}
\usepackage[utf8]{inputenc}
\usepackage[english]{babel}
\usepackage{graphicx}
\graphicspath{{Figures/}{../Figures/}}

\usepackage{amsmath}
\usepackage{graphicx}
\usepackage{caption}
\usepackage{subcaption}
\usepackage[export]{adjustbox}
\usepackage[position=b]{subcaption}

\usepackage{array}
\usepackage[none]{hyphenat}

\newcolumntype{x}[1]{%
>{\center\hspace{0pt}}p{#1}}%
\begin{document}
 
\begin{frontmatter}

\title{Computational design of organic solar cell active layer through genetic algorithm}


\author[add1]{Caine Ardayfio\corref{cor1}}
\ead{cardayfio@universityhighschool.org}

\cortext[cor1]{Please address correspondence to Caine Ardayfio}
\address[add1]{University High School, 2825 W 116th St, Carmel, IN, USA}

\begin{abstract}
The active layer microstructure of organic solar cells is critical to efficiency. By studying the photovoltaic properties of organic solar cell's microstructure, it is possible to increase the efficiency of the solar cell. A graph-based microstructure model was employed to approximate the efficiency, measured as short circuit current, of a solar cell given a microstructure. Through probabilistic graph-based optimization, a class of microstructures were found with an efficiency surpassing that of more conventional morphologies. These optimized solar cells surpass the efficiency of more conventional photovoltaic devices as they better facilitate charge transport, generation, and dissociation. A device was designed with a 40.29\% increase in short circuit current from the solar cells with the currently believed optimal morphology. The designed morphologies feature two dendritic clusters of the donor material poly(3-hexylthiophene-2,5-diyl) (P3HT) and the acceptor material phenyl-C61-Butyric-Acid-Methyl Ester (PCBM). The designed microstructure’s increase in performance contrasts with more conventional structures featuring interdigitated or bilayer strands of P3HT and PCBM. The change of microstructure morphology through graph-based evolution obtains an organic solar cell with an efficiency significantly greater than conventional organic solar cells, proves the validity of graph-based microstructure models for simulation in materials science, and advances the vision of an inexpensive, efficient form of renewable energy. 

\end{abstract}


\end{frontmatter}

\section{Introduction}
\label{sec1}

Organic solar cells (OSC) have acquired significant attention due to their low fabrication costs, simplicity in chemical modification (via chemical synthesis techniques), and their potential for large-scale manufacture. New fabrication techniques, such as roll-to-roll production, are enticing due to their production of inexpensive and light-weight photovoltaic devices \cite{hoth2012}. Alternative solar technologies, such as inorganic semiconductor devices, can only be produced through expensive processes like lithography or doctor blading, and do not achieve the ideal of inexpensive solar devices due to higher fabrication costs, especially when compared to the printing process in organic photovoltaics. As a result, organic solar cells are quickly becoming commercially viable photovoltaic devices.
 
 \begin{figure}[!t]
    \centering
    \includegraphics[width=0.6\linewidth]{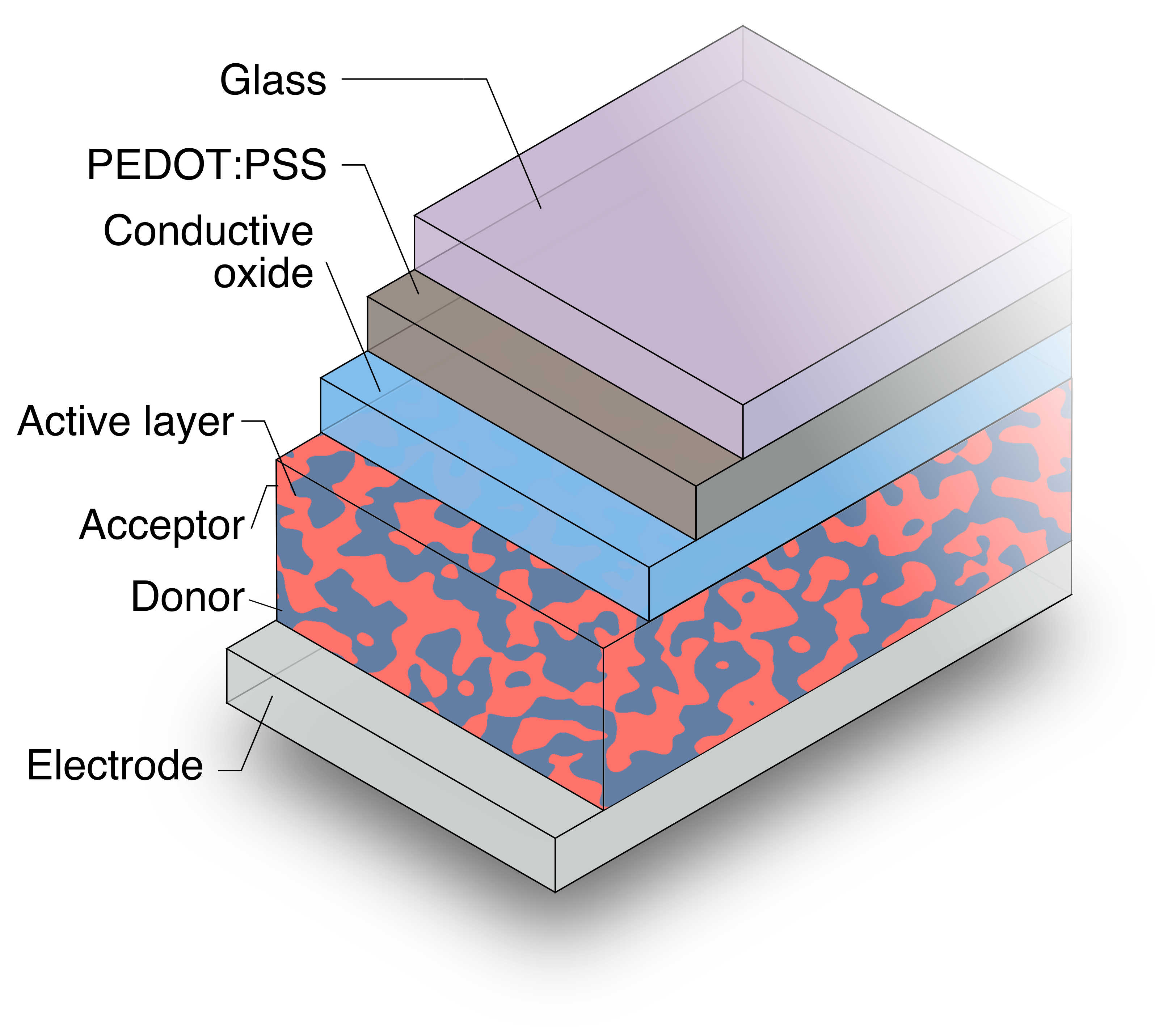}
    \caption{\textbf{Organic photovoltaic cell structure.} The chemical structure of modern organic solar cells, including: glass, to inhibit incident light reflection; active layer, for light absorption and exciton creation; and electrode, for exciton storage}
    \label{fig:Figure1}
 \end{figure}


However, the efficiency of OSCs is still lower than crystal or inorganic silicon solar cells. By studying the photovoltaic properties of organic solar cells, it is possible to increase the cell's efficiency. If the efficiency of organic solar devices were to be raised to just half that of inorganic solar devices, the production cost would be reduced to one-tenth the cost of inorganic solar devices per watt of power output. \cite{bagher2014}.

In the current development of solar cells, advancements have primarily been constrained to the chemical properties of the solar cells structure \cite{springer2007}. As a result, the efficiency of solar cells has been bottlenecked by morphology \cite{martens2003}.

Addressing this need, an evolution-inspired algorithm known as population-based incremental learning, was implemented to construct a more efficient, generic class of fractal morphologies that can be applied to a wide range of photoactive thin films. Various performance indicators derived from the exciton drift diffusion model \cite{wang2016} served as the objective function in the optimization of the solar cell. Derived from the exciton drift diffusion model, it was found that the microstructures transport of mass \cite{torquato2002}, charge \cite{salleo2010}, and reaction rate \cite{shearing2006} critically affect the efficiency of an organic solar cell.

By recognizing and optimizing these three properties, a microstructure was identified that exhibited efficiencies far greater than standard organic solar cells as evaluated by the drift diffusion equations.

\section{Results}
\label{sec2}

\subsection{Overview}
\label{subsec2.1}
A computational representation of the organic solar cell microstructure is presented. The microstructure model is graph-based with each vertex containing a label. The label denotes the probability of the material phase being a donor or an acceptor. The graph-based microstructure model is then fed through a genetic algorithm (population-based incremental learning) to incrementally improve the configuration of donor and acceptor material throughout the microstructure. The cost function utilized by the genetic algorithm is an approximation of efficiency with high-correlation to the physics-based efficiency (short-circuit current). This yields a solar cell with a donor-acceptor configuration that is significantly more efficient than standard configurations.

\subsection{Microstructure represented as graph}
\label{subsec2.2}
\begin{figure}[ht]
    \centering
    \includegraphics[width=1\linewidth]{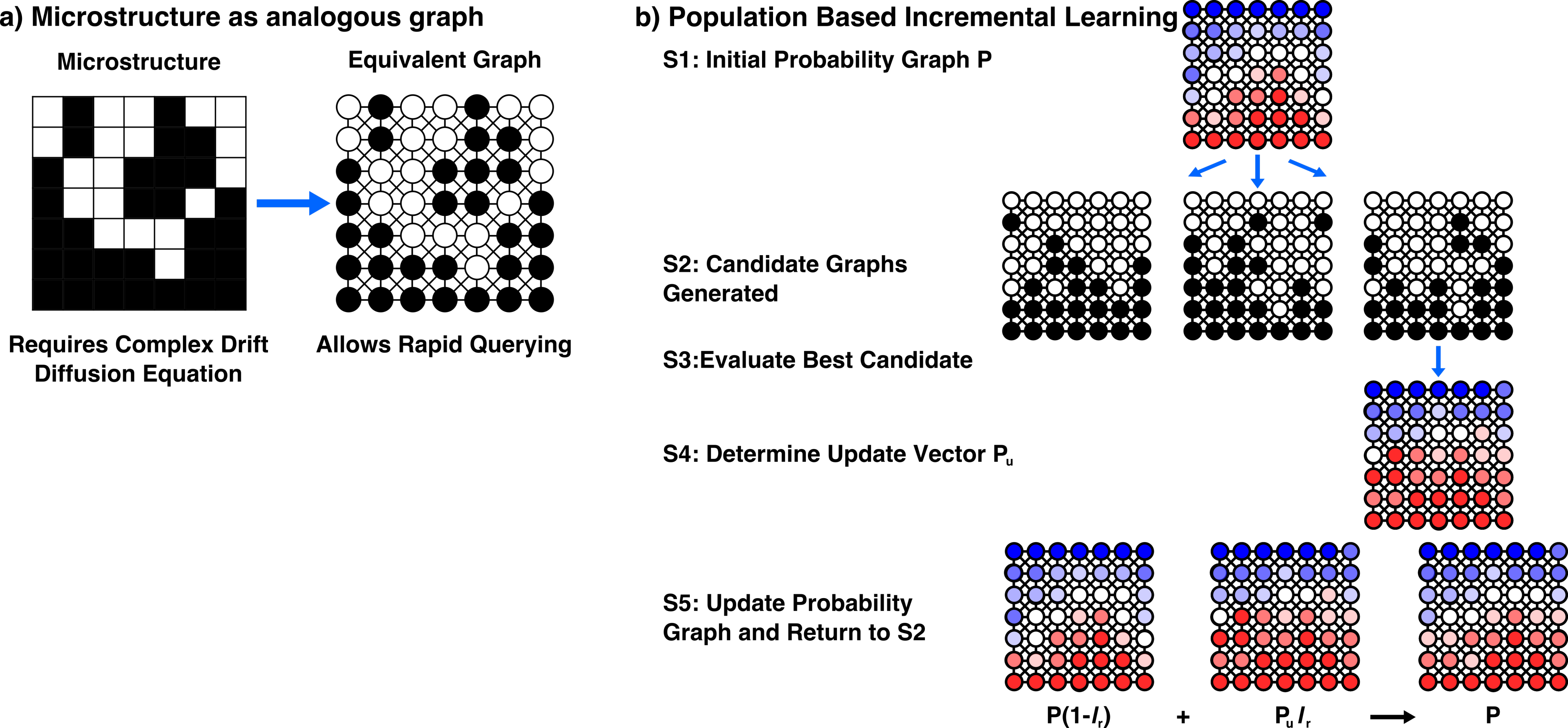}
    \caption{\textbf{Use of microstructure model as equivalent graph for evolutionary optimization.} \textbf{a} The concept of graphs as models for the microstructure present in solar cells allows efficient prototyping and evaluation of potential morphologies based off their respective efficiency equations; \textbf{b} Steps of PBIL: An initial probability distribution (S1) is sampled to create multiple morphologies (S2) which are quickly evaluated by efficiency (S3). The best candidate graph morphology will be translated to an update vector (S4) to be used in conjunction with the original graph P to create an updated probability graph (S5). Successive generations lead to probability distributions with better performance.}
    \label{fig:Figure2}
 \end{figure}

The efficiency of a solar cell microstructure through computational predictions involves interrogation of the morphology via a graph based microstructure model. The model maps the microstructure of the OSC to a two-phase graph. In the context of OSCs, the microstructure is the spatial configuration of donor and acceptor material in the solar cell. The microstructure model is constructed from the intrinsic parallels between a binarized matrix and a two-phase, donor-acceptor, morphology. The utility of a graph-based microstructure representation (see Figure \ref{fig:Figure2}) is the ability to exploit well studied algorithms from graph theory. 
These highly efficient algorithms have been instrumental for the optimization of many material models, including porous \cite{yu2008} and molecular microstructures \cite{ji2009, baluja1994}. The use of an equivalent graph to represent a microstructure will serve to streamline the construction of a cost function for optimization.

 Having constructed an equivalent graph from the microstructure, the graph is input through the probabilistic optimization algorithm that will evaluate the morphology ensuring specific properties are exhibited in the microstructure. 

\subsection{Population-Based Incremental Learning}
\label{subsec2.3}
Population-based incremental learning (PBIL), a type of probabilistic optimizer \cite{fairbanks2009}, was employed to design a highly efficient class of organic solar cells. Advantages of a probabilistic approach include the construction of a class of candidate microstructures, rather than a specific microstructure. PBIL also has an incredibly low memory foot print, making it accessible for use in non-specialized hardware. This does not hold for other forms of optimization such as deep neural networks. The general concept of the probabilistic algorithm is outlined in Figure \ref{fig:Figure2}. Each vertex in the probability vector denotes the probability of the vertex being an acceptor or donor phase, the chemical equivalents of PCBM and P3HT, respectively. The probability vector is updated every generation following evaluation by the cost function outlined below. The initial probability graph $\mathbf{P}$ begins with a vertex labeling vector (step S1 in Figure \ref{fig:Figure2}), which is a random distribution or a custom labeling vector. When a custom labeling vector was employed, the bilayer and interdigitated morphologies were represented with 75\% and 25\% values assigned to each pre-determined vertex in the explicit solution. This will ensure that the initial probability vector will initially tend towards a bilayer-like or interdigitated-like microstructure, but still allows divergence towards other morphologies. This exploits an advantage of PBIL, the ability to incorporate domain knowledge in the earliest stages of evolution. A number of candidate graphs are constructed from the existing probability distribution in each generation (step S2 in Figure \ref{fig:Figure2}). For all sampled morphologies, the cost function evaluates the performance and the best sample is identified (step S3 in Figure \ref{fig:Figure2}). Only the most efficient candidate graph’s probability distribution is used to calculate the update vector (step S4 in Figure \ref{fig:Figure2}). Specifically, the vector $\mathbf{P}_{\mathbf{u}}$ is an ‘update vector’ that describes a distribution of the labeled candidate graph performing best against the cost function. Every generation concludes by updating the distribution $\mathbf{P}$, $\mathbf{P}=\mathbf{P} \cdot\left(1-l_{r}\right)+\mathbf{P}_{\mathbf{u}} \cdot l_{r}$, where the rate of ‘learning’ is $l_{r}$. In each generation, the updated graph exhibits features deemed efficient in the morphology, and inhibits those features ill-suited towards the objective, at a constant learning rate $l_{r}$.  The learning rate, $l_{r}$, is set to 0.01 $<$ $l_{r}$ $<$ 0.4 but varies depending on the linear size of the graph. The iterations cease upon convergence to a single microstructure class. This is characterized when there is no improvement after 50 generations.

\subsection{Cost Function}
\label{subsec2.4}
The sole objective of the cost function $f$ is to optimize the following properties:

Absorption Efficiency $\mathrm{F}_{\mathrm{abs}}$: Fraction of light absorbing material

In a bulk heterojunction OSC, one material usually absorbs all incoming photons. As this is generally the donor material, a greater number of donors allows for more light absorption. However, light intensity decays with absorption depth as photons travel through a non-solid media from the top electrode. This decay is accounted for in the following: $\exp (-h(v) / H)$; where h is the physical distance from the vertex, v, to its respective electrode, and H is the absorption coefficient for the specific material.

Exciton Dissociation Efficiency $\mathrm{F}_{\mathrm{diss}}$: Fraction of photoactive material within the exciton diffusion length

When light is absorbed to generate excitons, the excitons must diffuse to an acceptor-donor interface where they can dissociate into separate charges. Excitons have a limited lifetime, in which there is a maximum distance, the exciton diffusion length, after which they recombine. At the point of recombination, the charge will often re-enter the lower energy band gap, eliminating any potential energy. The final $\mathrm{F}_{\mathrm{diss}}$ is calculated by determining the ratio of the light absorbing donors within the exciton diffusion length to the donor-acceptor interface.

Charge Transport Efficiency $\mathrm{F}_{\mathrm{out}}$: Fraction of corresponding paths to the respective electrode.

When an exciton has dissociated into its constituent parts, electron and hole, the path to the respective electrode must follow the respective domain (donor or acceptor). An optimal morphology minimizes the length of exciton transport paths for the produced charges to be transported through to contribute to the mass of usable current. This graph descriptor describes the ratio of the distance a charge must travel to reach the respective electrode through complementary phases over the distance indiscriminate of the phases and path taken. The minimum tortuosity, one, is exhibited in a solar cell when the shortest path through its respective phase to the respective electrode is equal to the shortest path, without consideration of phase, to the respective electrode. The tortuosity is determined for all available vertices in the graph and subsequently normalized to the data. The function punishes non-straight pathways that are susceptible to recombination.

\begin{figure}[ht]
\centering
\begin{subfigure}[b]{.5\textwidth}
  \centering
  \includegraphics[width=.76\linewidth,valign=t]{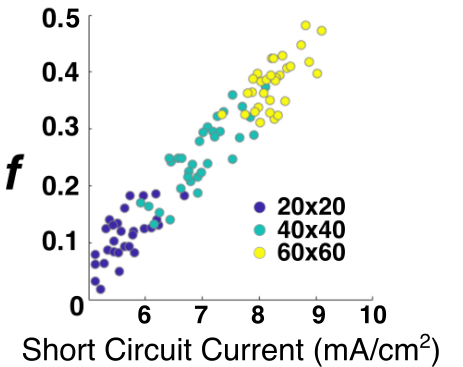}
  \caption{Correlation of approximated efficiency, $f$, and physics-based efficiency}
  \label{fig:Figure3a}
\end{subfigure}%
~
\begin{subfigure}[b]{.5\textwidth}
  \centering
  \includegraphics[width=.76\linewidth,valign=t]{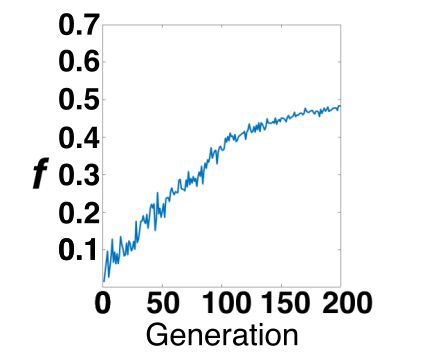}
  \caption{Solar cell performance over timesteps (generations) of PBIL}
  \label{fig:Figure3b}
\end{subfigure}
\caption{\textbf{Approximated efficiency of  model.} \textbf{a} The correlation between the device performance metric (short circuit current) and the approximated efficiency $f$, where $f$ is equal to $\mathrm{F}_{\mathrm{abs}}\mathrm{F}_{\mathrm{diss}}\mathrm{F}_{\mathrm{out}}$. The high correlation of the cost function with the physics-based efficiency ensures the evolutionary algorithm is optimizing the appropriate metric. \textbf{b} The evolution of the microstructures approximated efficiency over 200 generations of PBIL. This demonstrates the logarithmic nature of the efficiency for the most fit solar cell microstructure in each generation.}
\label{fig:Figure3}
\end{figure}

The product of these three property descriptors yields the physically apparent metrics for the efficiency of the specific microstructure. The metric plotted represents the product of $\mathrm{F}_{\mathrm{abs}}$, $\mathrm{F}_{\mathrm{diss}}$, and $\mathrm{F}_{\mathrm{out}}$ descriptors explained above and the physics-based short circuit current. Evaluation via these approximated efficiency graph descriptors provides a holistic performance rating that correlates well with the true microstructure efficiency found from the associated drift diffusion equations.

 \begin{figure}[ht]
    \centering
    \includegraphics[width=.6\linewidth]{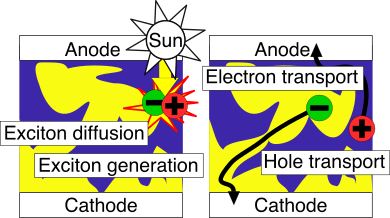}
    \caption{\textbf{Process of exciton generation, dissociation, and diffusion.} Energy conversion in OSCs begins with light absorption, in which a tightly bound exciton is created upon photon interaction with the donor material. An exciton is made up of a electron (anion) and hole (cation) which must be dissociated to generate current. This is done through exciton diffusion, where an exciton diffuses to the donor (P3HT) and acceptor (PCBM) interface. It is then split into its component parts through electrochemical forces. The component electron and hole then move to opposite electrodes}
    \label{fig:Figure4}
 \end{figure}
 
An analysis of the correlation between approximated efficiency and the physics-based short circuit current (see Figure \ref{fig:Figure3}) asserts the true efficiency from the associated drift diffusion equations has a high correlation with the approximated-efficiency. To understand how the property descriptors were derived to accurately approximate efficiency, an overview of the process of generation, dissociation, and diffusion for free carriers is presented (see Figure \ref{fig:Figure4}).

A bulk heterojunction OSC generates an exciton through light absorption of the active material, in this case, poly(3-hexylthiophene-2,5-diyl) (P3HT). Excitons are generated when a single electron is propelled from the highest occupied molecular orbit (HOMO) to the lowest unoccupied molecular orbit (LUMO), consequently, a hole is left in the HOMO. Note, in the case of OSCs, the energy differential between an electron in the LUMO and a hole in the HOMO is quite low compared to inorganic materials being just 1.3eV.  As a result, the Coulomb forces among the electrons are especially powerful making recombination highly likely for conventional microstructures. The photogenerated excitons will then be transported to the donor-acceptor interface to create an electron-hole pair, shortly after, the electron and hole diffuse to their respective electrode. The electron and hole are gathered by the electrodes and generate a current and voltage throughout the circuit-the product of the voltage and current is the total power the OSC produces. The dynamics of free carries incentivizes the maximization of the above property descriptors.

\begin{figure}[ht]
\centering
\begin{subfigure}[b]{.58\textwidth}
  \centering
  \includegraphics[width=.85\linewidth,valign=t]{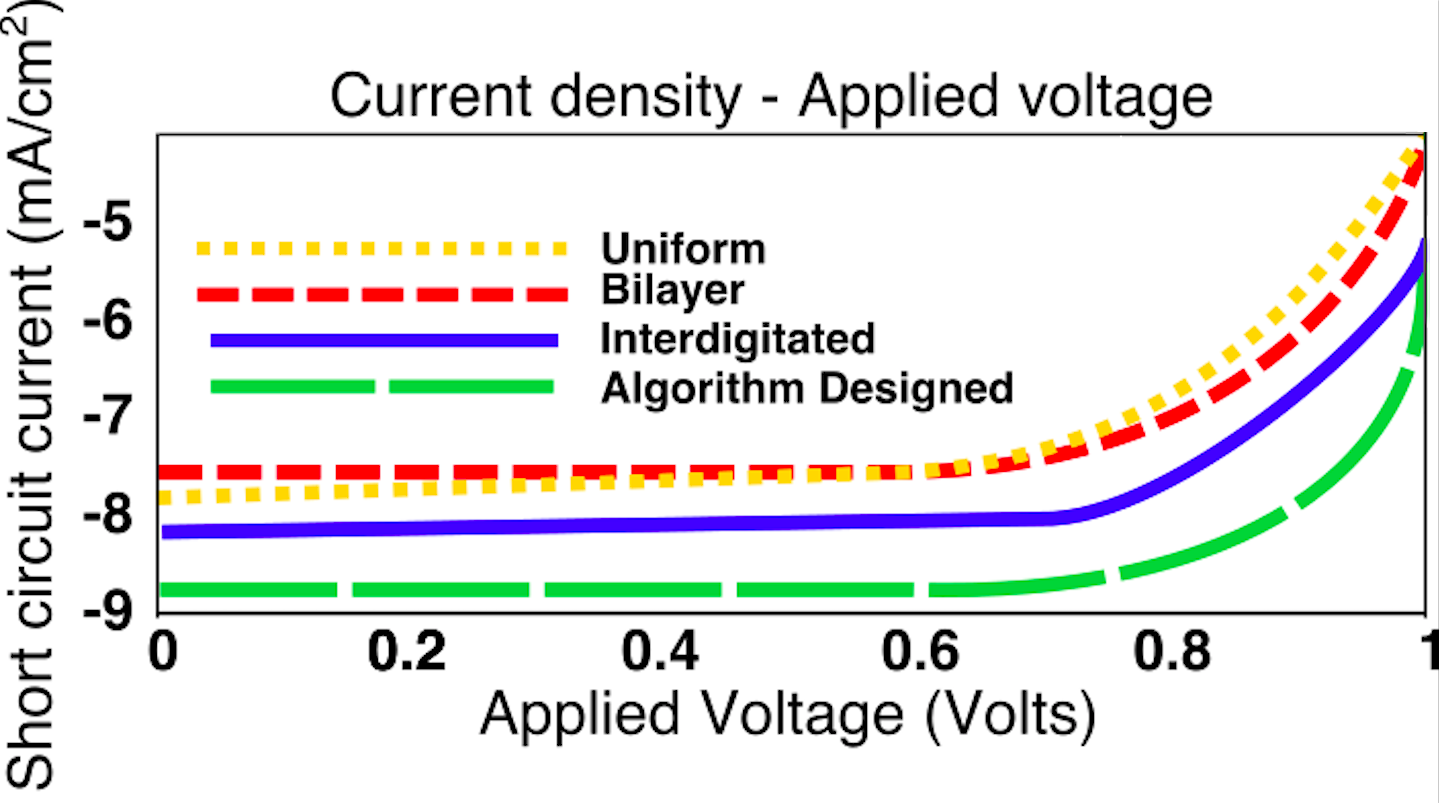}
  \caption{I-V Curve of designed versus typical microstructures }
  \label{fig:Figure5a}
\end{subfigure}%
~
\begin{subfigure}[b]{.42\textwidth}
  \centering
  \includegraphics[width=1\linewidth,valign=t]{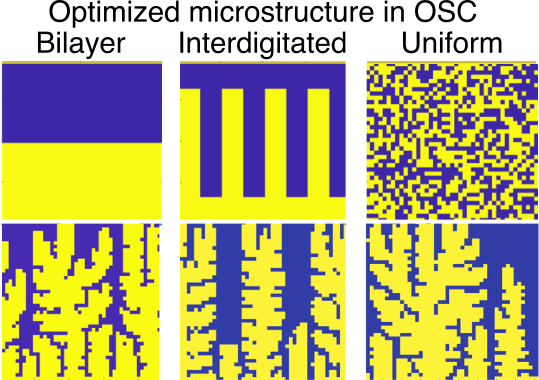}
  \caption{Optimized microstructure in OSC}
  \label{fig:Figure5b}
\end{subfigure}
    
\caption{\textbf{Resultant microstructure efficiency.} \textbf{a} The energy-voltage curves for uniform, bilayer, interdigitated, and the designed morphologies. These curves were evaluated with the exciton drift-diffusion model \cite{wang2016}. \textbf{b} The morphogenesis from initially bilayer, interdigitated, and uniform distributions towards the final optimized structure. Different initial probability distributions (Top) evolve towards dendritic morphologies with similar properties (Bottom), notably, fractal dimension. Purple denotes the donor material (P3HT) and yellow denotes the acceptor material (PBCM).}
\label{fig:Figure5}
\end{figure}

\begin{table}\centering
\begin{tabular}{|l|c|c|c|}\hline
                      & \multicolumn{3}{c|}{Microstructure Evolution for Different Initial Distributions}               \tabularnewline\hline 
                      & \multicolumn{1}{c|}{Bilayer} & \multicolumn{1}{c|}{Interdigitated} & \multicolumn{1}{c|}{Uniform} \tabularnewline\hline
Short circuit current   & 8.36 & 8.81 & 8.42\tabularnewline\hline
Approximated efficiency & 0.40 & 0.48 & 0.41\tabularnewline\hline
Fractal dimension       & 1.79 $\pm$ .07 & 1.72 $\pm$ .12 & 1.71 $\pm$\ .06\tabularnewline\hline
\end{tabular}
\caption{Approximated efficiency, $f$, fractal dimension, $D_{f}$, and short circuit current, $J_{sc}$ ($mA/cm^{2}$) of designed microstructures from initially bilayer, interdigitated, and uniform distributions.}
\label{table:Table1}
\end{table}

Zi Shuai Wang et al., having proven the validity of the drift-diffusion model \cite{wang2016}, a reliable method for numerical simulation of organic solar cells, and an efficient, graph-based approximation were extracted from the parameters. The governing equations for the free carriers have 3 main parameters, all of which have been incorporated into the previously mentioned Q1, Q2, and Q3 of the approximated cost function $f$.

Having obtained a graph microstructure model, cost functional $f$, and probabilistic graph-based optimization method, an evolved morphology was found.

The transformation of the morphology from a starting uniform, bilayer, and interdigitated microstructure is presented in Figure \ref{fig:Figure5}. A photovoltaic thickness of 100nm and homogeneous light absorption is assumed as that is typical for OSC devices \cite{he2011,lu2015}. The probabilistic optimization designed a dendritic structure, which is not used in conventional organic solar cell design and shows great performance increases from previously thought optimal morphologies. The I-V (current-voltage) curves of the numerical simulations for various solar cell morphologies, including the designed morphology, are shown in Figure \ref{fig:Figure5}. The designed microstructure exhibits an elegant combination of three seemingly disparate constraints, a donor-acceptor interface lengthy enough to enable efficient exciton dissociation, a large amount of donor material to maximize light capture, and fully connect domains to enable efficient exciton transfer. An interdigitated morphology is assumed to be the optimal microstructure for precisely this reason. However, the use of sub-optimal, man-made, morphologies is completely avoided with the use of PBIL and mathematical cost functions that produce more efficient morphologies. Interestingly, the optimal morphologies exhibit fractal characteristics which will be explored later.

The predicted short circuit current, $\mathrm{J}_{\mathrm{sc}}$, for the most efficient of the designed microstructures optimized through the PBIL is 8.81$mA/cm^{2}$ (see Table \ref{table:Table1}), 40.29\% more efficient than the short circuit current of 6.28$mA/cm^{2}$ exhibited by the more conventional interdigitated columnar morphology \cite{shaw2008}. This is likely due to the microstructure facilitating greater exciton dissociation with insignificant increases in recombination. An increased length of the donor–acceptor interface is also seen that resulted in better exciton dissociation while maintaining a relatively non-tortuous charge transport pathway. This served to inhibit performance decreases due to exciton recombination.

\subsection{Integration of domain knowledge}
\label{subsec2.5}
Due to the nature of PBIL, domain knowledge is easily incorporated into the development of the microstructure. As opposed to starting from an initially uniform distribution, the morphogenesis began with a good initial guess, the interdigitated morphology, and a bad initial guess, the bilayer. The use of previously gained knowledge on optimal microstructures as starting points in the morphogenesis proved to be very effective as the initial guess tending towards the interdigitated morphology proved to be about 4.63\% more efficient than the morphology evolved from an initially random guess.

\subsection{Fractal Analysis}
\label{subsec2.6}
The dendritic-fractal microstructure arising from the evolution of the morphology exhibits obvious fractal properties. Fractal analysis was employed to find that the fractal dimension of the optimal microstructure was 1.76, similar to that of many space filling curves \cite{mccarthy2008}. Fractal analysis was utilized to confirm the validity of the designed morphologies through similarity comparison of naturally occurring fractals.

 \begin{figure}[h!]
    \centering
    \includegraphics[width=0.8\linewidth]{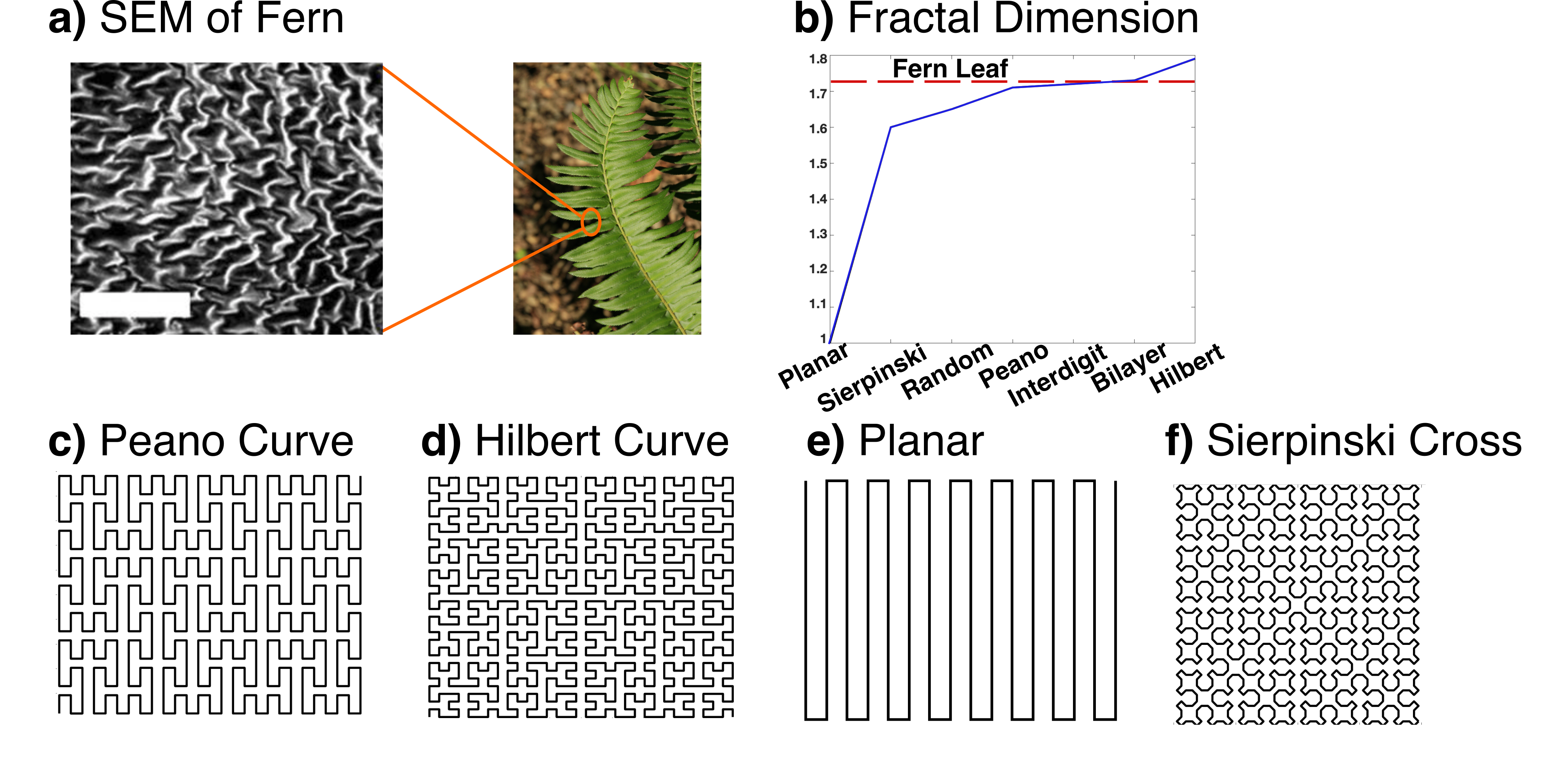}
    \caption{\textbf{Fractal comparison of space-filling curves with designed morphology.} \textbf{a} SEM of fern leaf (Polystichum munitum). \textbf{b} Calculated Hausdorff Fractal Dimension for c-f, OSC structures evolved from initially bilayer, random, and interdigitated conditions, and fern leaf. Scale bar 100 $\mu$m \textbf{c} Peano space-filling curve, \textbf{d} Hilbert space filling curve, \textbf{e} planar structure, and \textbf{f} Sierpinski cross}
    \label{fig:Figure6}
 \end{figure}

Fractal analysis methods such as fractal dimension \cite{shearing2006}, lacunarity \cite{sumpter1994}, and succolarity \cite{johnson2018}, can be used to infer the specialized functions of a structure such as the movement of electricity through the fractal dendrites in neurons and the transport of particles in the bronchi of the lungs \cite{szarko2014}. In this case, fractal analysis was used to confirm the validity of the solar cell microstructure in transporting and generating excitons. 

The designed microstructure has parallels to the internal makeup of Polystichum munitum (see Figure \ref{fig:Figure6}), generally known as the Barnsley fern. Among the literature, it is firmly established that the leaves of Polystichum munitum, and more general ferns, feature high efficiency for the storage of energy in biological process.
The internal structure and transportation network of the ferns bares resemblance, statistical and visual, to fractal space-filling curves (see Figure \ref{fig:Figure6}). The high-energy storage in fern leaves due to a large surface area to volume ratio is analogous to the production and transport of excitons in a solar cell. Both structures must create a surface with maximal area to optimize photon absorption while also maintaining a large area for the particle dissociation. By analyzing fractals similar to the fern leaf, and comparing them to the designed solar cell microstructure, we can gain insight into the degree of optimization present in the solar cell.

Many space-filling curves are compared, the Peano curve, the Hilbert curve, the Sierpinski Cross, and a planar structure as a control. The available storage area in the family of space filling curves is found by the evaluation of the curves Hausdorff dimension, calculated by the box-counting algorithm \cite{onsager1938,braun1984}. The relationship between Hausdorff dimension, the fractal S, and the box side-length $\varepsilon$ can be defined with the following.
\begin{equation}
\operatorname{dim}_{\mathrm{box}}(S) :=\lim _{\varepsilon \rightarrow 0} \frac{\log N(\varepsilon)}{\log (1 / \varepsilon)}
\operatorname{dim}_{\mathrm{box}}
\end{equation}
 
Using MATLAB, the fractal dimensions were determined, and it was found that the Hilbert and Peano fractal had the highest fractal dimension of 1.73, close to the dimension of the fern leaf. These space-filling curves serve as inspiration for nature-inspired surfaces in solar cells due to their fractal dimension similar to that of the designed microstructure morphology and naturally occurring fractals.

\begin{figure}[h!]
    \centering
    \includegraphics[width=0.8\linewidth]{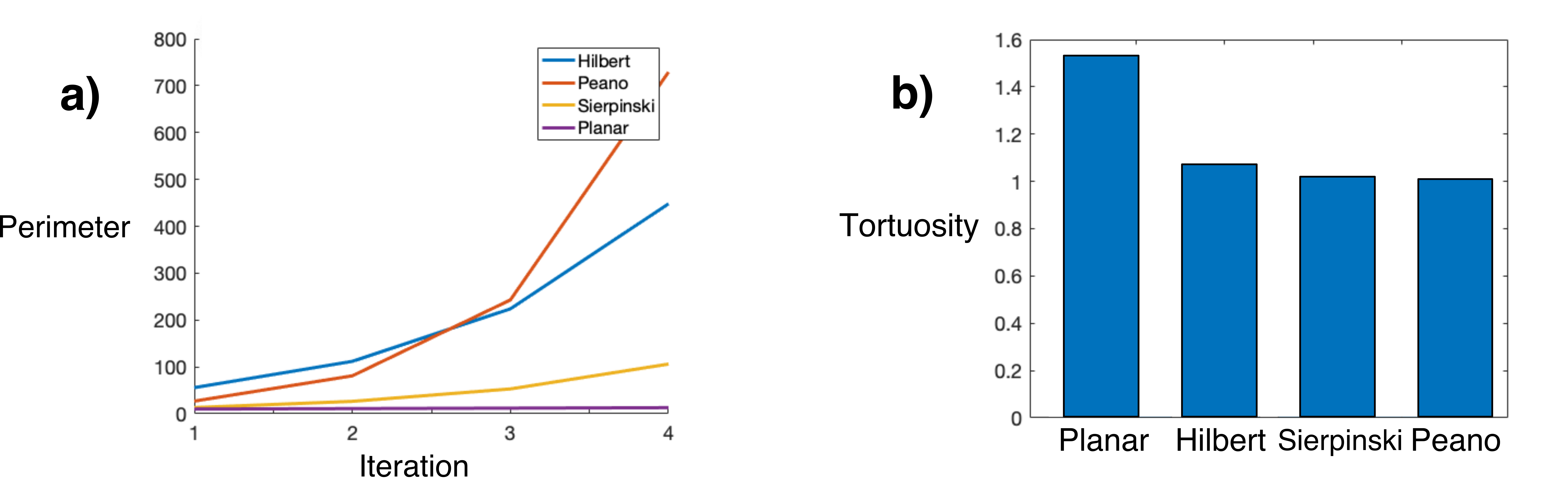}
    \caption{\textbf{Perimeter and tortuosity calculated for different space filling curves.} \textbf{a} Length of fractals sampled over various recursion depths. The fractal with the highest length would serve as a model of donor-acceptor interface; \textbf{b} Tortuosity of fractal curves. Calculated by the mean distance to the end of the structure with barriers divided by the distance to the end of the structure without barriers.}
    \label{fig:Figure7}
 \end{figure}

The efficiency of OSCs can be improved through the following mechanisms. The first mechanism is based on the length of interface between the donor and acceptor. The second mechanism is the tortuosity of the path towards the respective electrode. The third mechanism is the number of donor materials absorbing incident light. These mechanisms were explained in further detail previously as the basis for the objective function for the PBIL. The Hilbert and Peano fractal, both with a fractal dimension similar to the fern leaf and designed morphology, appear to optimize these three mechanisms the most. Figure \ref{fig:Figure7} shows the Peano and Hilbert curves have the highest perimeter, a mechanism especially helpful in interface design. In addition, Figure \ref{fig:Figure7} shows the tortuosity of the fractals. This was calculated in a similar manner to the $\mathrm{F}_{\mathrm{out}}$ method, in which the distance to the end of the structure with barriers is compared to the distance to the end of the structure without barriers. Again, the fractal curves showed promise for facilitating fast transport due to their non-tortuous nature. Applying naturally-occurring fractal curves to solar devices allows optimization of properties charge transport and exciton dissociation, allowing more efficient solar cells.

\subsection{Conclusion}
\label{subsec2.7}
Microstructure design is a crucial part of developing organic solar cells. OSCs have the potential to become ubiquitous amongst power generation due to their ease of fabrication and low price. Although there have been incremental improvements in the chemical properties of solar cells, a lack of progress in finding the optimal morphology has greatly inhibited organic solar cell adoption. In this, it is illustrated how high-performance microstructures can be developed rapidly via a graph-based strategy. This is in stark contrast to the trial-and-error methods currently employed for organic solar cell microstructure optimization. Treating the microstructure of a material system as graphs allows modular and extensible models that are simple to query and evaluate. The graph  model quickly maps the microstructures properties and integrates well with optimization algorithms while elegantly integrating prior domain knowledge into the microstructure design process. This use of graph-based modeling and probabilistic optimization results in a microstructure design with a 40.29\% higher efficiency than conventional solar designs. Fractal analysis was also used to further prove the validity of the designed morphologies. This was accomplished through analyzing models analogous to the function of the solar cell and comparing their similarity with the designed fractal structure. To conclude, graph-based probabilistic optimization led to the identification of a class of microstructures that feature significantly higher efficiencies than currently leading solar cells. It is anticipated this method, coupled with fractal analysis, will be useful for microstructure optimization and implementation.

\section{Methods}
\label{sec3}

All code was created in MATLAB and is freely available online \cite{ardayfio2019}. This includes the implementation of the population based incremental learning and the cost function used to evaluate the realizations. In each generation of the microstructure optimization, 160 candidate graph morphologies were generated. There were 500 generations allocated to optimization, but due to early convergence towards an optimal morphology, optimization ceased continuation after 200 generations. All realizations were evaluated with the cost function explained above and the most efficient microstructures were used to compute the update vector. The rate of learning was set to 0.1. The cost function was created to have high correlation with the drift diffusion equations governing efficiency for a two-phased (donor-acceptor) organic solar cell. The complete model, unlike the microstructure model, evaluates the drift diffusion equation and determines the short circuit current $J_{sc}$. This evaluation was applied solely to the final evolved microstructures to significantly decrease optimization time and to prove the utility of a graph-based microstructure model. The fractal dimension of all microstructures were calculated to mathematically evaluate statistical similarity.

\section{Acknowledgements}
\label{sec4}
Declarations of interest: none

This research did not receive any specific grant from funding agencies in the public, commercial, or not-for-profit sectors.

This work was supported by Brandon and Meredith Hogan, Kathleen Armato, and University High School of Indiana.

\clearpage

\bibliographystyle{elsarticle-num}
\bibliography{references}

\end{document}